\documentclass[sigconf, screen]{acmart}
\AtBeginDocument{%
  }
\setcopyright{acmlicensed}
\copyrightyear{2018}
\acmYear{2018}
\acmDOI{XXXXXXX.XXXXXXX}
\acmConference[MM '25]{Proceedings of the 33rd ACM International Conference on Multimedia (ACM MM)}{October 27--31,
  2025}{Dublin, Ireland}
\acmISBN{978-1-4503-XXXX-X/2018/06}

\usepackage{array}
\usepackage{makecell}
\usepackage{multirow}
\usepackage{pifont}
\usepackage{bm}
\usepackage{bbm}
\usepackage{tikz}

\usepackage[ruled,vlined]{algorithm2e}
\newcommand{\cmark}{\ding{52}}%
\def\eg{{\em e.g.}}

\begin{document}
\renewcommand\footnotetextcopyrightpermission[1]{}
\title{Bridging Vision and Language: Optimal Transport-Driven Radiology Report Generation via LLMs}

\author{Haifeng Zhao}
\orcid{0000-0002-5300-0683}
\email{senith@163.com}
\affiliation{%
        \department{School of Computer Science and Technology}
	\institution{Anhui University}
	\city{Hefei}
	\state{Anhui}
	\country{China}
}

\author{Yufei Zhang}
\orcid{0009-0003-3699-7703}
\email{summoning@stu.ahu.edu.cn}
\affiliation{%
        \department{School of Computer Science and Technology}
	\institution{Anhui University}
	\city{Hefei}
	\state{Anhui}
	\country{China}
}

\author{Leilei Ma}
\orcid{0000-0001-8681-0765}
\email{xiaomylei@163.com}
\affiliation{%
        \department{School of Computer Science and Technology}
	\institution{Anhui University}
	\city{Hefei}
	\state{Anhui}
	\country{China}
}

\author{Shuo Xu}
\orcid{0009-0008-6655-9401}
\email{xush1020@163.com}
\affiliation{%
        \department{School of Computer Science and Technology}
	\institution{Anhui University}
	\city{Hefei}
	\state{Anhui}
	\country{China}
}

\author{Dengdi Sun}
\orcid{0000-0002-0164-7944}
\email{sundengdi@163.com}
\affiliation{%
        \department{School of Artificial Intelligence}
	\institution{Anhui University}
	\city{Hefei}
	\state{Anhui}
	\country{China}
}

\renewcommand{\shortauthors}{Haifeng Zhao et al.}


\begin{abstract}
Radiology report generation represents a significant application within medical AI, and has achieved impressive results. Concurrently, large language models (LLMs) have demonstrated remarkable performance across various domains. 
However, empirical validation indicates that general LLMs tend to focus more on linguistic fluency rather than clinical effectiveness, and lack the ability to effectively capture the relationship between X-ray images and their corresponding texts, thus resulting in poor clinical practicability.
To address these challenges, we propose Optimal Transport-Driven Radiology Report Generation (OTDRG), a novel framework that leverages Optimal Transport (OT) to align image features with disease labels extracted from reports, effectively bridging the cross-modal gap. 
The core component of OTDRG is Alignment \& Fine-Tuning, where OT utilizes results from the encoding of label features and image visual features to minimize cross-modal distances, then integrating image and text features for LLMs fine-tuning. 
Additionally, we design a novel disease prediction module to predict disease labels contained in X-ray images during validation and testing.
Evaluated on the MIMIC-CXR and IU X-Ray datasets, OTDRG achieves state-of-the-art performance in both natural language generation (NLG) and clinical efficacy (CE) metrics, delivering reports that are not only linguistically coherent but also clinically accurate. 

\end{abstract}








\maketitle

\section{Introduction}
Radiology imaging is a cornerstone of pulmonary disease diagnosis, yet the escalating workload of radiologists threatens its efficacy. 
Radiologists process considerable quantity X-ray images daily, with misdiagnosis rates reaching up to 3\%-5\% in high-demand settings due to time constraints and fatigue~\cite{first}. 
This gap underscores the urgent need for automated tools to bridge the divide between images and free-text descriptions~\cite{2016cvpr,tienet}, generating accurate, contextually rich radiology reports. Therefore, the radiology report generation has attracted the attention of numerous researchers in recent years and become an important application of AI.

\begin{figure}[t]
\centering
  \includegraphics[width=1\linewidth]{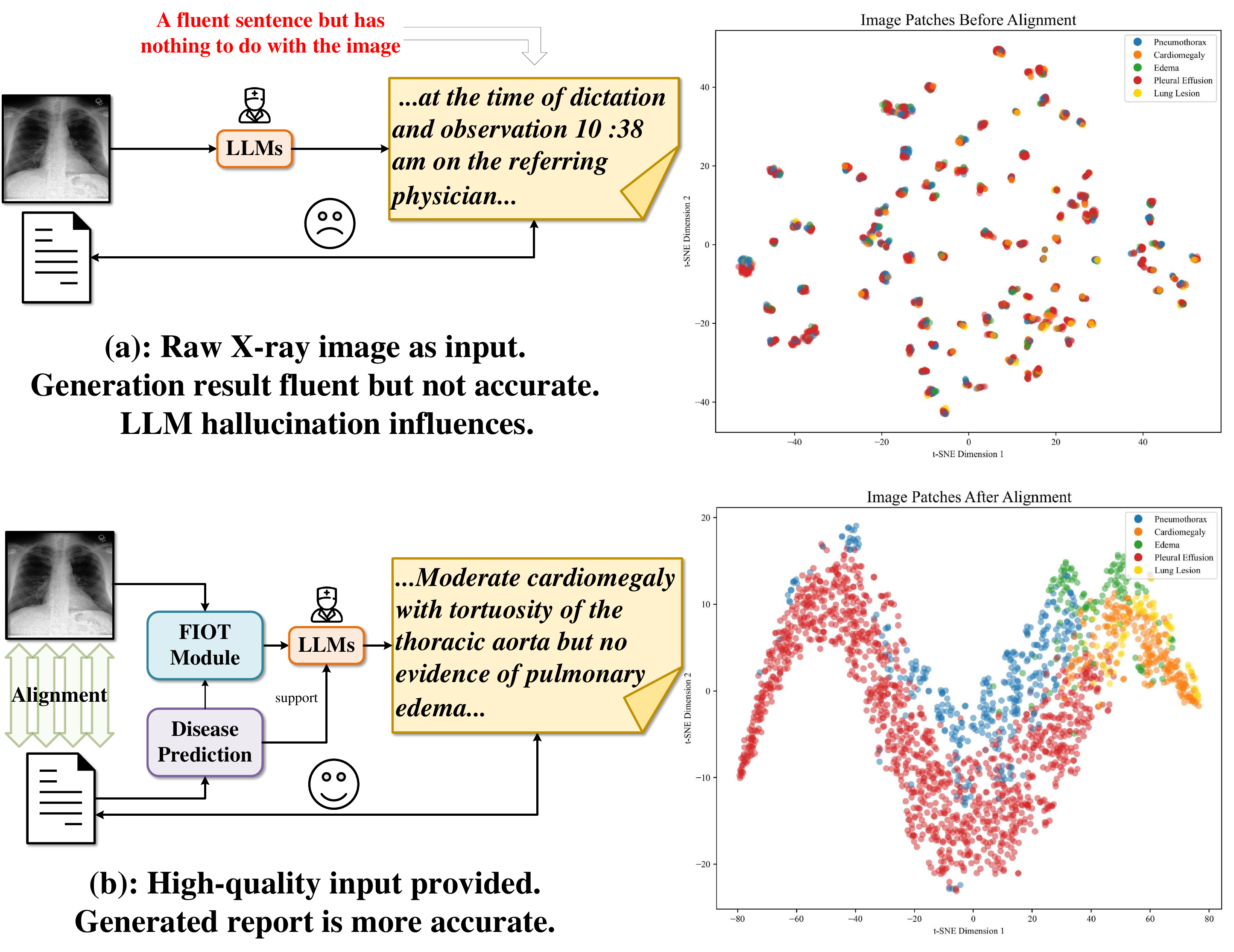}
\caption{The example in the figure briefly illustrates the effectiveness of our method. The left figure shows the difference between the methods, and the right figure shows the corresponding image feature distribution.}
\label{fig:ff}  
\end{figure}

Although traditional methods have made great progress in the task of radiology report generation, they struggle with generalization and reliability~\cite{r2gen,metransformer,kiut}, often producing reports with poor vocabulary diversity and structural coherence.
Recently, the advent of large language models (LLMs) has sparked hope for utilizing their unparalleled text generation capabilities in the task~\cite{qwen,llama2,gpt4}, and thus also holds great potential in generating radiology reports. 
However, the highly specialized nature of medicine imposes higher requirements for LLMs, and current radiology report generation methods based on LLMs still fail to deliver satisfactory results, as there remains a discernible gap between the radiological reports they generate and actual clinical diagnoses. To summarize, it currently confronts the following two key challenges:


\textbf{\textit{Challenge} 1: LLMs may prioritize fluency over clinical accuracy.} When applied to radiology report generation, LLM-based methods such as R2GenGPT~\cite{r2gengpt} exhibit a critical limitation: while generating linguistically fluent reports, their performance remains mediocre from a clinical evaluation perspective~\cite{xraygpt}. This deficiency stems from the inherent constraints of language-exclusive LLMs, whose probabilistic pattern-prediction mechanisms~\cite{llmmechanism} disproportionately prioritize textual fluency over diagnostic precision. Such technical shortcomings typically manifest persistent hallucinations, contextual logical discontinuities, insufficient domain-specific sensitivity, etc., seriously degrading the quality of generated radiology reports~\cite{llmhallucination1,llmhallucination2}. In addition, in the field of medicine, the complexity of professional knowledge and the limitations of model training data will further exacerbate this problem. As shown in Figure~\ref{fig:ff}(a), the generated report reveals the LLM-induced hallucinatory content that, although present in the training corpus, fundamentally lacks relevance to the corresponding medical imagery.



\textbf{\textit{Challenge} 2: Unaligned image-text inputs make it difficult for LLMs to extract meaningful information.} 
Radiology report generation requires models to have a deep understanding of specialized terminology and clinical contexts. 
Low-quality input, such as X-ray images themselves or unprocessed reports, may not provide sufficient context for the model to accurately comprehend and generate appropriate medical content.
Consequently, the crux of employing LLMs for radiology report generation lies in providing more medically-focused input.
The common strategy for enabling the text generator to understand X-ray images is cross-modal alignment. 
Mainstream methods in aligning X-ray image-report pairs include memory-driven methods~\cite{r2gencmn}, knowledge graph-based methods~\cite{gsket}, contrastive learning methods~\cite{CLR2G}, etc.~\cite{DCL, boundingbox}.
However, these alignment methods often underutilize the complex nonlinear relationships of feature distribution and cannot capture the global structure, which is not good enough for models to bridge the modal gap. 
The two images on the right of Figure~\ref{fig:ff} show the data distribution before and after alignment using our method, where each color corresponds to a disease. 
After the features are aligned, the image regions belonging to different diseases are clearly separated, and the generated reports are more clinically relevant.

To overcome these challenges and unlock the full potential of LLMs in this domain, a more robust and powerful alignment method is needed. 
In recent years, Optimal Transport (OT) has demonstrated its advantages in many computer vision tasks like object detection~\cite{otobject}, knowledge distillation~\cite{otdistillation}, etc.~\cite {otprompt}, which motivates our introduction of OT in the radiology report generation task.
In our model, OT functions as an innovative cross-modal alignment strategy, establishing a global connection between X-ray images and the medical knowledge encapsulated within their corresponding reports, thereby significantly enhancing the accuracy and relevance of the generated radiology reports.
The efficacy of OT in our approach stems from its ability to mathematically align the distributions of X-ray image features and textual disease labels, addressing the inherent modality gap that limits LLM performance in vision-language tasks.
This global optimization, enhanced by the Sinkhorn-Knopp algorithm with entropic regularization~\cite{sinkhorn,wasserstein}, ensures stable and efficient convergence while preserving semantic correspondence.

In summary, we propose OTDRG (\textbf{O}ptimal \textbf{T}ransport-\textbf{D}riven \textbf{R}adiology Report \textbf{G}eneration), a model designed to enhance the clinical efficacy of generated radiology reports.
\textbf{Firstly,} to enable LLMs to obtain richer medical knowledge, we draw on prior works~\cite{CheXbert,promptmrg} and utilize a frozen LLM to extract disease labels from reports, classifying them into 14 primary groups with 4 distinct states.
These labels are regarded as auxiliary information and represent the fundamental knowledge within diagnostic reports.
\textbf{Secondly, }by segmenting X-ray images into patches and tokenizing the disease labels, we frame the task of minimizing cross-modal distances as an OT problem. 
This strategy aims to bridge the gap between medical images and paired reports through optimization, quantifying the cross-modal distance using a cost matrix and iteratively identifying the most aligned image representations with the text.
\textbf{Thirdly,} outside the model training workflow, we have also designed a disease label recognition method that involves converting predicted and true disease labels into one-hot encoding for learning, enabling the model to recognize the disease status contained in radiology images more comprehensively.
The resulting vectorized images, disease labels, reports, and prompts are then used to train the LLM, ultimately improving the clinical efficacy of the generated radiology diagnostic reports. The generated reports demonstrate state-of-the-art performance on both natural language generation and clinical evaluation metrics.

In conclusion, our contributions can be summarized as follows:
\begin{itemize}
\item We propose the utilization of OT to bridge the modal gap between images and their corresponding reports for alignment, thereby enabling LLMs to comprehend the correlation between them more effectively.
\item We design a novel model to generate clinically accurate radiology reports utilizing LLMs, which considers a combination of aligned images, disease labels, reports, and prompts as input.
\item We develop an innovative method to predict disease labels effectively from the X-ray images, which can be regarded as summaries of the reports, supporting LLMs to generate radiology reports.
\end{itemize}

\begin{figure*}[t]
\centering
  \includegraphics[width=1\linewidth]{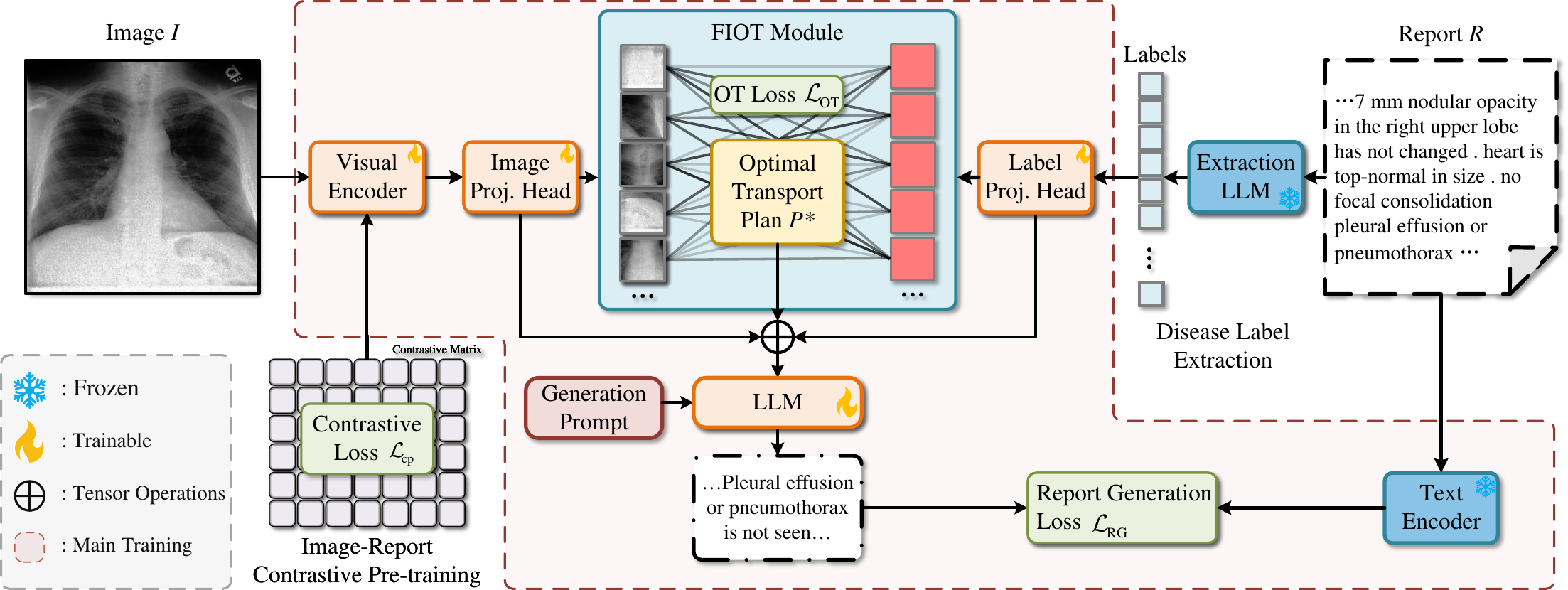}
\caption{The overall workflow of OTDRG, which includes a visual encoder, a text encoder, an image projection head, and a label projection head. The training process contains two preparation steps, which are Disease Label Extraction and Image-Report Contrastive Pre-training, and the main training process focuses on cross-modal alignment and fine-tuning.}
\label{fig:framework}  
\end{figure*}

\section{Related Work}
\subsection{Radiology Report Generation}
The generation of radiology reports has recently emerged as a prominent topic in the medical field, particularly as an application of multimodal tasks. 
Traditionally, researchers have utilized models based on Convolutional Neural Networks (CNN) or Transformer, but the models achieved high performance mostly by following traditional text generators (\eg, BERT)~\cite{CNN,attention,bert}.
Extensive research has made significant progress in this field.
~\cite{r2gen} and ~\cite{r2gencmn} focused on shared memory to capture the internal alignment of paired image and text. 
Clinical-BERT aimed to improve the ability to comprehend medical knowledge by pre-training a vision-language model with a large set of clinical text~\cite{clinicalbert}. 
Many works attempted to integrate knowledge modules into radiology report generation tasks, extracting the key knowledge to support generating reports with greater concern on clinical quality~\cite{gsket,knowledgematters}.
METransformer proposed multiple learnable expert tokens diving into different image regions to learn complementary representations using both linear and non-linear attentions~\cite{metransformer}. 
PromptMRG presented a framework with the prompt of diagnostic results to produce medically correct reports, with self-adaptive disease-balanced learning to enhance the ability to classify rare diseases in the dataset, overcoming the barrier of the text decoder's inability to manipulate disease distributions~\cite{promptmrg}. 
Previous methods have made significant contributions to the field, but they are limited by the traditional text decoder. Our model incorporates LLMs to harness robust text processing capabilities for radiology report generation.
\subsection{Optimal Transport}
Optimal Transport (OT) is a mathematical theory and a computational method that describes the distance or correspondence between two probability distributions~\cite{monge}. 
Its core concept is how to best transfer a set of resources from one location to another. 
The first application that was widely disseminated of OT in computer vision is Earth Movers' Distance, as a way to evaluate dissimilarity between two multi-dimensional distributions in feature space~\cite{earth}.
Researchers found that OT enhances performances in various fields in deep learning, like domain adaptation~\cite {otdomain}, object detection~\cite{otobject}, knowledge distillation~\cite{otdistillation}, semantic correspondence~\cite{otsemantic}, and prompt learning~\cite{otprompt}. 
Our work suppose OT to align X-ray image features and the disease label features, in order to bridge the cross-modal gap.
\subsection{Large Language Model}
Large Language Models (LLMs) have made significant advancements in the field of Natural Language Processing, which focus on processing and generating textual data. 
By training on large-scale text corpora, they have learned to understand the structure, semantics, and context of the input text, performing exceptionally well in various tasks.  
For instance, models like LLaMA~\cite{llama2}, Qwen~\cite{qwen}, and GPT~\cite{gpt4} not only continuously redefine our understanding of machines' capabilities to comprehend and generate human language on a technical level but also demonstrate immense potential and value in practical applications. 
LLMs are fine-tuned in our model with Low-Rank Adaptation (LoRA)~\cite{lora} strategy on medical cross-modal data, enabling it to handle downstream tasks in the medical field and process image-text tasks related to medicine with fewer trainable parameters.

\section{Methodology}
\subsection{Problem Definition and Model Overview}
In the radiology report generation task, our goal is to generate the most accurate report for X-ray image diagnosis.
We define the $N$ original X-ray images as $I = \{i_1, i_2, \cdots, i_n\}$ and the corresponding $N$ diagnosis reports as $R = \{r_1, r_2, \cdots, r_n\}$, and follow the existing work and regard $R$ as ground truth to make the report $R^* = \{r^*_1, r^*_2, \cdots, r^*_n\}$ that generated by LLMs as similar as possible to the ground truth.
We treat the cross-modal alignment of the image feature and disease label feature as an Optimal Transport problem, modeling the cross-modal distance in a common LLM's feature space as $d_{\text{OT}}$, and aim at find the optimal transport plan matrix $T^*$ in $t$ times of iteration to minimize $d_{\text{OT}}$, achieving the gap-bridging.

As illustrated in Figure \ref{fig:framework}, the core component of OTDRG is alignment and fine-tuning; in addition, it also includes the encoding of label features and image visual features. Specifically, a frozen LLM extracts key information from ground truth reports $R$ to generate $N$ disease labels $L = \{l_1, l_2, \cdots, l_n\}$ (Sec.~\ref{sec:dle}), and the Image-Report Contrastive Pre-training (IRCP) strategy is used to enhance the ability of visual encoder to capture the visual features of X-ray images (Sec.~\ref{sec:ircp}). In the OT strategy, the above visual feature $X = f_e(I)$ and disease label feature $Y = f_e(L)$ are further projected onto the LLMs feature space by different trainable projection heads. Thus $f_p(X)$ and $f_p(Y)$ are generated for the Feature Integration by Optimal Transport (FIOT) module.
Then, the aligned features will be combined with the generation prompt and disease label itself to form a high-quality input, allowing LLMs to better understand the medical information contained in X-ray images and their relationship with diagnostic reports (Sec.~\ref{sec:fiot}). 
For the fine-tuning of LLMs, we adopt the LoRA strategy and use supervised fine-tuning to fine tune LLMs by comparing the diagnostic reports generated by LLMs based on the above inputs with the original reports (Sec.~\ref{sec:finetune}).
Outside of the training process of OTDRG, we have designed a disease prediction module that can predict disease labels contained in X-ray images during validation and testing, replacing disease labels extracted from truth reports during training (Sec.~\ref{sec:prediction}).

\subsection{Preliminary}
\subsubsection{Disease Label Extraction}
\label{sec:dle}
Inspired by previous work~\cite{CheXbert}, we describe the X-ray image by 14 diseases, which are \textit{enlarged cardiomediastinum, cardiomegaly, lung opacity, lung lesion, edema, consolidation, pneumonia, atelectasis, pneumothorax, pleural effusion, pleural other, fracture, support devices, no finding}. Then, 4 states are introduced to represent each disease as \textit{positive, negative, unclear, not mentioned in the report}, we define disease labels with the form of \texttt{disease}: \texttt{state}.
The combination of disease and states as labels can almost describe all situations in the diagnosis report corresponding to images, and the labeled representation is more advantageous in cross-modal feature alignment because they have already extracted semantic information from semantically rich and complex reports.

We extract disease labels by employing a frozen LLM with low parameters, and we employ a meticulously designed prompt to augment the frozen LLM's capability to infer disease labels, which will enhance both natural language generation and clinical perception of the generated reports by our model.

\subsubsection{Visual Feature Extraction via IRCP}
\label{sec:ircp}
In order to enable the visual encoder in OTDRG to better process X-ray images, we use contrastive learning and pre-train at the image-report level.
Following the mainstream method to extract features for medical images, we choose the Swin Transformer~\cite{swin} as our visual encoder, pre-trained with paired medical images and texts for better understanding of image inputs.
Formally, the visual feature extraction process can be denoted as:
\begin{align}
    f_e(I) = \bm{X} = \{\bm{x}_1, \bm{x}_2, \ldots, \bm{x}_{N_{s}}\}~,
\end{align}
where \(\bm{x}_i \in \mathbb{R}^{D}\) is a patch embedding, \(D\) denotes the feature dimension, and \(N_{s}\) is the number of patches.

To perform contrastive learning, we generate positive samples by adding Gaussian noise to the extracted image features and treat other samples in the same batch as negative samples. Specifically, for a selected anchor feature \(\mathbf{a}\), we create a positive key \(\mathbf{a}^+\) by:
\[
\mathbf{a}^+ = \mathbf{a} + \mathcal{N}(0, \sigma^2 \mathbf{I})~,
\]
where \(\sigma = 0.1\) is the standard deviation of the Gaussian noise.

The IRCP loss is based on the InfoNCE loss~\cite{infonce}, which encourages the model to maximize the similarity between the anchor and its positive key while minimizing the similarity with negative samples. The loss for a single anchor is formulated as:
\begin{equation}
\mathcal{L}_{\text{IRCP}} = -\log \frac{\exp(\text{sim}(\mathbf{a}, \mathbf{a}^+) / \tau)}{\exp(\text{sim}(\mathbf{a}, \mathbf{a}^+) / \tau) + \sum_{\mathbf{k} \in \mathcal{K}} \exp(\text{sim}(\mathbf{a}, \mathbf{k}) / \tau)}~,
\end{equation}
where \(\text{sim}(\cdot, \cdot)\) is the similarity function (\eg, Euclid Distance), \(\tau = 0.5\) is the temperature parameter, and \(\mathcal{K}\) represents the set of negative keys, typically other samples in the batch.

\subsection{Alignment and Fine-tuning}
\label{sec:fiot}
LLMs as tools for text processing are not supposed to take image features as input, which exhibit a distribution distinctly different from text features in the projection space.
After we get the visual features and disease labels, we need to think about treating these two features to better serve the LLMs to generate radiology reports.
In that case, we introduce Feature Integration by Optimal Transport (FIOT), a novel strategy using OT to minimize the gap between image features and text features in the projection space, presenting the image features to be analyzed in a more textual manner.
\subsubsection{Feature Projection}
Before we integrate the multimodal features, we are supposed to project the image and disease label features into the same space. 
The extracted diseased labels are advised to be encoded for disease features formally as
\begin{align}
f_e(L) = \bm{Y} = \{\bm{y}_1, \bm{y}_2, \cdots, \bm{y}_{n}\}~,
\end{align}
where $\bm{y}_i \in \mathbb{S}$, $i \in \{1, 2, \cdots, n\}$, denoting each disease label, $\mathbb{S}$ is the disease label set.
Afterward, the visual feature $f_e(I)$ and disease features $f_e(L)$ will be mapped onto the LLaMA input feature space, where we are going to employ OT for further processing. The projection can be presented formally as:
\begin{align}
    \text{$f_p(X)$} & = \text{Proj}(X) = \mathbf{W_i} \times \text{$f_e(I)$} + \mathbf{b_i}~,\\
    \text{$f_p(Y)$} & = \text{Proj}(Y) = \mathbf{W_l} \times \text{$f_e(L)$} + \mathbf{b_l}~,
\end{align}
$f_p(I)$ and $f_p(L)$ represent the mapped image batch feature and disease label feature, $\mathbf{W_i}$ and $\mathbf{W_l}$ are weight matrix in the process of linear transformation, and the $\mathbf{b_i}$ and $\mathbf{b_l}$ are bias terms when the projection process happens.
\subsubsection{OT for Bridging the Modal Gap}
When both radiology image features and disease label features that extracted from matching reports are projected on the shared feature space, these two types of features exhibit different characteristic distributions. 
Although large language models are supposed to take pure text input for best performance, this poses a significant challenge in multimodal tasks, especially in radiology report generation.
During the research, we notice that Optimal Transport (OT) can be an outstanding solution in various deep learning tasks, which inspired us to utilize it on our report generation task.
OT considers the difference between image features and disease label features in a mathematical aspect, measuring the cross-modal distance, and finding the closest image presentation textually through optimization iterations. 

In the radiology report generation background, a discrete situation is more related to the problem setting.
We define $\mathbf{C}$ as the cost matrix for OT, which is formulated as:
\begin{align}
    \mathbf{C}_{i,j} = \sqrt{\sum\nolimits_{i=1,j=1}^{M,N} (f_p(X)_i - f_p(Y)_j)^2}~.
\end{align}
The matrix $\mathbf{C}$ represents the Euclidean distance between every point in $f_p(X)$ and $f_p(Y)$, denoting the transportation cost in the OT. 
The goal for our model is to minimize the cross-modal distance $d_\text{OT}$ by an optimal transport plan $T$, and formally, the problem can be written as:
\begin{equation}\label{eq:ot}
\begin{aligned}
    d_{\text{OT}}(u, v|\mathbf{C}) & = \underset{T}{\text{minimize}}<T, \mathbf{C}>\\
    \qquad \text{s.t.} \quad T1_N & = u,  T^\mathrm{T}1_M = v,
\end{aligned}
\end{equation}
where $u, v$ are two discrete probability vectors initialized so that they sum to 1, and during the calculation, the elements in $u, v$ denote the probability mass to each point.
However, optimizing roughly may burden computational resources and be time-consuming, which is why we utilize the Sinkhorn-Knopp algorithm~\cite{sinkhorn}.
The Sinkhorn-Knopp algorithm boasts fast computational speed and high numerical stability, making it suitable for use in large-scale cross-modal datasets.
To solve the OT problem more rapidly, we need to use an entropic constraint for the Sinkhorn-Knopp algorithm, then the optimization problem can be formulated as:
\begin{equation}\label{eq:ot}
\begin{gathered}
    d_{\text{OT},\varepsilon}(u, v|\mathbf{C}) = \underset{T}{\text{minimize}} \langle T, \mathbf{C} \rangle - \varepsilon \Omega(T) \\
    \text{s.t.} \quad T1_N = u,~T^\mathrm{T}1_M = v, ~ T \geq 0, 
    \Omega(T) = \sum\nolimits_{i,j} T_{i,j} \log(T_{i,j})~,
\end{gathered}
\end{equation}
where $\Omega(T)$ denotes the entropy, and $\varepsilon$ is the hyper-parameter, and both of them aim to find smoother and more stable transportation plans while promoting rapid convergence of the algorithm. By adjusting the weight of entropy, a balance can be struck between transportation costs and the uniformity of the transportation plan.
More details and the formula for solving the OT problem with the Sinkhorn-Knopp algorithm in FIOT are presented in Algorithm 1.

During the OT process, the optimal transportation plan $T^*$ can be found to minimize the total cost (including the entropy regularization term).
This leads to finding representations in the common embedding space that bring the image and report, considering as disease labels,  modalities closer together, thereby bridging the modal gap, allowing for a more nuanced understanding of the relationship between visual findings and diagnostic criteria that can be better understood by LLMs. Here, we present the workflow of the FIOT algorithm as algorithm~\ref{alg:sinkhorn}.
\definecolor{codeblue}{rgb}{0.25,0.5,0.5}
\newcommand\mycommfont[1]{\footnotesize\ttfamily\textcolor{codeblue}{#1}}
\SetCommentSty{mycommfont}
\begin{algorithm}[!ht]
\small
\DontPrintSemicolon
\SetNoFillComment
\textbf{Input:} Image Feature $X$, Disease Label Feature $Y$, uniform marginal $\boldsymbol{\mu}$ for row constraint , $\boldsymbol{\nu}$ for column constraint,\\
\textbf{Parameter:}  Regularization Parameter $\varepsilon$, Iteration Times $t$,\\
\textbf{Output:} Minimal Cross-modal Distance $d_\text{OT}$\\ 
$f_p(I)\xleftarrow{\stackrel{\text{Proj}}{}}X, $
$f_p(L)\xleftarrow{\stackrel{\text{Proj}}{}}Y$\\
$C \gets \text{Euclid Distance}(f_p(X), f_p(Y))$\\
$K = \text{exp}^{-C/\varepsilon}$ \\
$d_I, d_L \gets \text{Shape of Dimension 0 of  } f_p(X), f_p(Y)$\\
$\bm{\alpha} \gets \text{OnesInit}(d_I),  \bm{\beta}\gets \text{OnesInit}(d_L)$\\
$\bm{\alpha}_i \gets \bm{\alpha}_i/(d_I), \bm{\beta} \gets \bm{\beta}_i/(d_L)$\\
\While{$i < t$}{
$\bm{\alpha}^t = \boldsymbol{\mu} / (K\bm{\beta}^{t-1})$\\
$\bm{\beta}^t = \boldsymbol{\nu} / (K^T\bm{\alpha}^t)$\\
$\bm{\alpha}, \bm{\beta} = \text{softmax}(\bm{\alpha}, \bm{\beta})$\\
}
$\text{Transport Plan  } T^* = \text{diag}(\bm{\alpha})~K~\text{diag}(\bm{\beta})$\\
\textbf{return} $d_\text{OT} = \underset{T}{\text{minimize}}<T, \mathbf{C}> - h(T^*)$
\caption{FIOT with Sinkhorn-Knopp algorithm}
\label{alg:sinkhorn}
\end{algorithm}

\subsubsection{Fine-Tuning Details}
\label{sec:finetune}
In OTDRG, LLMs are chosen as our report generator. The original LLMs have limited capability in handling medical knowledge, so our model's overall objective is to endow LLMs with the ability to diagnose X-ray images and generate radiology reports through supervised fine-tuning.
For the prompt of LLMs, we connected $f_p(I)$ and $f_p(L)$ to achieve the cross-modal gap-bridging, enriching the input with more medical prior knowledge and more semantic information.
As we discussed before, LLMs prefer accepting inputs of textual modality, and via the FIOT module, the raw X-ray images are made to gravitate towards textual modality in terms of feature representation.
The feature-integrated prompt $F$ is designated as 
\begin{flushleft}
\texttt{</IMG> $f_p(I) \bigoplus f_p(L)$ </IMG>}\\
\texttt{Generate a comprehensive diagnosis report for this radiology image. \#\#\#Focus on diagnostic accuracy\#\#\#}
\end{flushleft} 
By framing the input in this way, we leverage the natural language processing capabilities of LLMs to understand the correlation between the visual findings from the medical images and the diagnostic criteria. 
It not only enhances the LLMs' ability to generate coherent and contextually relevant medical reports but also ensures that the output is aligned with the standard practices of diagnosis.

Previous research has found that LLMs do not perform well when parameters are frozen. However, setting all parameters to be trainable statements leads to unaffordable computational resource usage and time expenses.
During the process of model training, we introduced the LoRA strategy to fine-tune the LLMs~\cite{lora}. 
Introducing two trainable low-rank matrices $\mathbf{W}_a$ and $\mathbf{W}_b$ to approximate the weight changes $\Delta \mathbf{W}$ in the full-parameter fine-tuning of LLMs, thereby significantly reducing the number of parameters required for fine-tuning and lowering the model training costs.
For LLMs' weight matrix $W$, LoRA restricts its updates in the following manner of $\mathbf{W}^{'} = \mathbf{W} + \Delta \mathbf{W} = \mathbf{W} + \mathbf{W}_{a} \mathbf{W}_{b}$.

\begin{figure}[t]
\centering
  \includegraphics[width=1\linewidth]{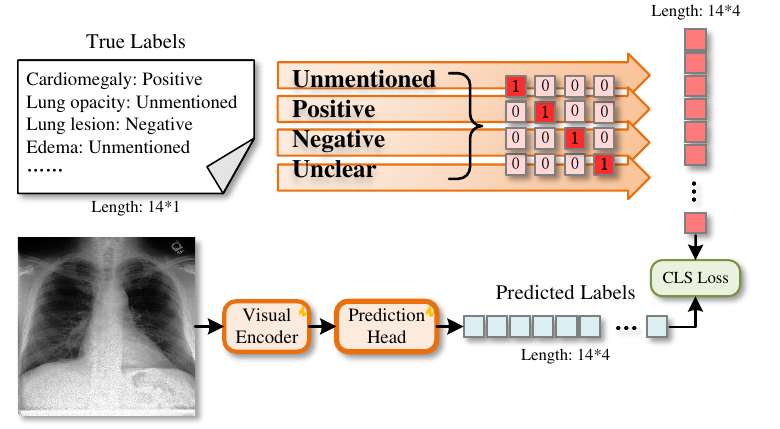}
\caption{Workflow of the disease label prediction.}
\label{fig:cls}    
\end{figure}

\subsection{Disease Label Prediction}
\label{sec:prediction}
During our research, we notice that accurate extraction of disease status labels is essential (more details can be found in the Supplementary). 
As mentioned in the DLE section, fundamentally, we view a disease label as \texttt{disease}: \texttt{state}.
We further transform labels into one-hot encoding that is conducive to model learning by abstracting the four states of a disease into an array of length 4:
\begin{equation}\label{eq:0-1}
\begin{aligned}
    \quad l_{i} &=
    \begin{cases}
    [1,0,0,0], & \text{when unmentioned} \\
    [0,1,0,0], & \text{when positive} \\
    [0,0,1,0], & \text{when negative} \\
    [0,0,0,1], & \text{when unclear}
    \end{cases},
\end{aligned}
\end{equation}
where $l_i$ is the $i$-th \texttt{disease}: \texttt{state} label.

\begin{table*}[t]
\centering
\caption{\small Comparison with SOTA radiology report generation methods on MIMIC-CXR dataset. $\#$ indicates the usage of LLMs for generation. All results are from the original paper. The best results are in \textbf{bold}, and the second-best results are \underline{underlined}.}
\label{tab:mimiccxr}
\begin{tabular}{lllccccccc}
\toprule	 
\multirow{2}{*}{\textbf{Dataset}} & \multirow{2}{*}{\textbf{Model}} & \multirow{2}{*}{\textbf{Year}} & \multicolumn{3}{c}{\textbf{CE metrics}} & \multicolumn{4}{c}{\textbf{NLG metrics}}\\ \cmidrule(r){4-6} \cmidrule(r){7-10}
 & & &Precision & Recall & F1 & BLEU-1 & BLEU-4 & METEOR & ROUGE-L \\
\midrule
\multirow{11}{*}{\makecell{\textbf{MIMIC-CXR}}} & R2Gen & 2020 & 0.333 & 0.273 & 0.276 & 0.353 & 0.103 & 0.142 & 0.277 \\
& CliBert & 2022 & 0.397 & 0.435 & 0.415 & 0.383 & 0.106 & 0.144 & 0.275 \\
& M2KT & 2023 & 0.420 & 0.339 & 0.352 & 0.386 & 0.111 & - & 0.274 \\
& METrans. & 2023 & 0.364 & 0.309 & 0.311 & 0.386 & 0.124 & 0.152 & \underline{0.291} \\
& KiUT & 2023 & 0.371 & 0.318 & 0.321 & 0.393 & 0.113 & 0.160 & 0.285 \\
& DCL & 2023 & 0.471 & 0.352 & 0.373 & - & 0.109 & 0.150 & 0.284 \\
& R2GenGPT$^{\#}$ & 2023 & - & - & - & 0.399 & \underline{0.126} & 0.160 & 0.286 \\
& CLR2G & 2024 & 0.398 & 0.302 & 0.319 & 0.388 & \underline{0.126} & 0.158 & 0.286 \\
& PromptMRG & 2024 & \textbf{0.501} & \textbf{0.509} & \textbf{0.476} & \underline{0.398} & 0.112 & 0.157 & 0.268 \\\
& MCSAM$^{\#}$ & 2024 & - & - & - & 0.379 & 0.109 & 0.149 & 0.284 \\
& AdaMatch$^\#$ & 2024 & - & - & - & 0.379 & 0.106 & \underline{0.163} & 0.286\\
\cmidrule(r){2-10}
& \textbf{Ours} & - & \underline{0.474} & \underline{0.448} & \underline{0.436} & \textbf{0.411} & \textbf{0.128} & \textbf{0.164} & \textbf{0.295} \\
\bottomrule
\end{tabular}
\end{table*}

In the report generation process, we divide this vector into subgroups of 4 elements each, representing each disease, and then take the maximum prediction probability value of each subgroup as the label for that disease in the report generation.
Outside of the model's training workflow, we additionally train a small classification model that can learn the aforementioned vector to extract disease labels from original medical images, as shown in Figure~\ref{fig:cls}.
Because the predicted labels after 0-1 transactions are sparse and there is an issue of class distribution imbalance, we introduce Focal loss~\cite{focal} to compare the differences between true labels and predicted labels, thereby improving the model's precision and recall in predicting labels, and it can be formulated as:
\begin{equation}\label{eq:focal}
\begin{aligned}
    \mathcal{L}_\text{cls} &= \frac{1}{N} \sum\nolimits_{n=1}^{N} \left[ \left( 1 - \exp \left( -\ell(l_i,p_i) \right) \right)^\gamma 
    \cdot \ell(l_i,p_i)  \right],\\
    &\mathrm{s.t.}\quad\ell(l_i,p_i) = -\sum\nolimits_{i=1}^{C} l_i \log(p_i) \\
\end{aligned}
\end{equation}
where $p_i$ is the $i$-th prediction label, $l_i$ denotes the $i$-th one-hot encoding of the true label, $N$ and $C$ represent the number of samples and the length of the prediction vector, and $\gamma$ is a hyper-parameter.

\subsection{Learning Objectives}
In this section, we define the learning objectives of the model, including the loss for generating medical reports and the optimal transport loss.

\subsubsection{Medical Report Generation Loss} In our work, cross-entropy is employed to meticulously assess and compare the generated reports by fine-tuned LLMs against the reports in the ground truth within our model. 
This choice is strategically driven by cross-entropy's proven merits, particularly its adeptness at quantifying the divergence between the prediction probability distribution and the actual distribution of the ground truth, offering a reliable metric for the model's predictive accuracy in the context of medical diagnostics.
Our model are trained to maximize $p_{\theta}(y_t|y_1, \ldots,y_{t-1},F)$ by minimizing the following:
\begin{align}
   \mathcal{L}_\text{RG} = -\sum\nolimits_{t=1}^{T}\log p_{\theta}(y_t|y_1, \ldots,y_{t-1},F)
\end{align}
where $y_t \in (y_1, y_2, \ldots, y_T)$ denotes LLM's prediction token at step $t$, $\theta$ is the trainable parameter, $F$ represents the feature-integrated prompt mentioned before. 
\subsubsection{Optimal Transport Loss}
As an essential part of the model, the FIOT module returns the cross-modal Sinkhorn distance $d_\text{OT}$ between the images and disease labels.

$d_\text{OT}$ is calculated by the transport plan. Thus, the goal in training is to find the optimal transport plan by the Sinkhorn-Knopp algorithm and calculate the minimum $d_\text{OT}$ for generating better clinically-wise diagnostic reports.
During the model training process, we treat $d_\text{OT}$ calculation formula as the $\mathcal{L}_\text{OT}$ function, thereby endowing the medical image can be presented as textual features as LLMs' input.
Our models can be stimulated to perform better because the input features of LLMs are more akin to textual form.

The overall learning objective can be defined formally as:
\begin{align}
   \mathcal{L} = \mathcal{L}_\text{RG} + \lambda\mathcal{L}_\text{OT}~,
\end{align}
where $\lambda$ is a trade-off parameter.

\section{Experiments}
\subsection{Datasets}
\paragraph{IU X-Ray} IU X-Ray~\cite{iuxray}~\footnote{https://openi.nlm.nih.gov/} is a widely used dataset in radiology report generation field, which contains 2,955 paired medical image-report samples after preparation in advance. Evaluation with this dataset is frequently employed in recent research.
\paragraph{MIMIC-CXR} The MIMIC-CXR dataset~\footnote{https://physionet.org/content/mimic-cxr/}, as introduced by prior work~\cite{mimic}, comprises an extensive collection of radiology photographs accompanied by corresponding reports, representing the most substantial repository of its kind in the field of medical radiography. Adhering to the standard dataset division and pre-processing techniques outlined by ~\cite{CheXbert}, the refined dataset is organized into 270,790 instances for training, 2,130 for validation, and 3,858 for testing purposes.

\subsection{Evaluation metrics}
\paragraph{NLG metrics} Natural language generation (NLG) metrics evaluate the quality generated reports using algorithms automatically. NLG metrics in our research includes BLEU~\cite{bleu}, ROGUE-L~\cite{rouge}, and METEOR~\cite{meteor} to comprehensively assess the reports.
\paragraph{CE metrics} We introduce Clinical Efficacy (CE) metrics following works ~\cite{ce}, which includes precision, recall, and F1 metrics evaluate reports by converting reports into 14 disease labels using CheXbert~\cite{CheXbert}, representing the generated reports' performance on clinical scenarios.
As mentioned earlier, due to the radiology reports generated by LLMs perform poorly on CE metrics. 
Changing this situation is a key consideration when designing our model.

\subsection{Model Analysis}

\subsubsection{Comparison Study}

\begin{table}[t]
\caption{\small Comparison experiments on IU X-Ray dataset. $\#$ indicates the usage of LLMs for generation. The best results are in \textbf{bold}, and the second-best results are underlined.}
\label{tab:iu}
\centering
\begin{tabular}{lcccccc}
\toprule	 
\multirow{2}{*}{\textbf{Model}} & \multicolumn{4}{c}{\textbf{NLG metrics}}\\ \cmidrule(r){2-5}
 & BLEU-1 & BLEU-4 & METEOR & ROUGE-L \\
\midrule
R2Gen & 0.325 & 0.059 & 0.131 & 0.253 \\
M2KT & 0.371 & 0.078 & 0.153 & 0.261 \\
DCL & 0.354 & 0.074 & 0.152 & 0.267 \\
R2GenGPT$^\#$ & 0.488 & 0.173 & 0.211 & \textbf{0.438} \\
PromptMRG & 0.401 & 0.098 & 0.160 & 0.281 \\
CLR2Gen & 0.493 & \textbf{0.184} & 0.202 & 0.378 \\
MCSAM$^\#$ & 0.489 & \textbf{0.184} & 0.210 & \underline{0.394} \\
\cmidrule(r){1-5}
\textbf{Ours} & \textbf{0.501} & \underline{0.182} & \textbf{0.215} & 0.377 \\
\bottomrule
\end{tabular}
\end{table}

To evaluate the performance of OTDRG, we conduct a comprehensive comparison with several SOTA radiology report generation methods on the MIMIC-CXR and IU X-Ray datasets.
We consider R2GenGPT~\cite{r2gengpt} as our baseline model for its innovation in using LLMs for radiology report generation. 

\textbf{Comparisons on MIMIC-CXR Dataset} In Table~\ref{tab:mimiccxr}, we compare OTDRG with other SOTA radiology report generation methods on MIMIC-CXR dataset, including R2Gen~\cite{r2gen}, ~\cite{M2KT}, ClinicalBert~\cite{clinicalbert}, METransformer~\cite{metransformer}, KiUT~\cite{kiut}, DCL~\cite{DCL}, R2GenGPT, AdaMatch~\cite{adamatch}, CLR2G~\cite{CLR2G}, PromptMRG~\cite{promptmrg} and MCSAM~\cite{mcsam}. 
The results demonstrate that OTDRG achieves SOTA results, especially in NLG metrics, where our models all achieve the highest scores and were second only to PromptMRG in CE metrics. Notably, PromptMRG's performance in the NLG metrics is poor, even failing to rank among the top three in the most important BLEU-4 metric. Furthermore, we surpass all models based on LLMs. Therefore, in summary, our approach has superior overall performance in terms of language fluency and clinical accuracy.

\textbf{Comparisons on IU X-ray Dataset} In Table~\ref{tab:iu}, we measure our model on the IU X-ray dataset, and clearly, our model performs well on this dataset.
It is worth noting that CE metrics are not suitable for the IU X-Ray dataset.
The lower ROUGE-L score of OTDRG on the IU X-Ray dataset may be because its cross-modal alignment and language style fail to fully adapt to the heterogeneity and non-standardized reporting characteristics of the dataset.
Furthermore, the MIMIC-CXR dataset not only has a vast amount of data but also follows the same \textit{14 diseases} pattern in previous work and ours, which makes the DLE module more accurate in describing the data, which explains why our model can achieve impressive performance on the MIMIC-CXR dataset in the textual aspect.

In addition, our findings highlight OTDRG’s effectiveness in leveraging OT for cross-modal alignment and LLM fine-tuning, significantly enhancing clinical accuracy over methods like R2GenGPT, which prioritize textual fluency over clinical efficacy. 
Comprehensively, the superior NLG and CE scores on MIMIC-CXR, combined with competitive results on IU X-Ray, underscore the proposed OTDRG’s robustness and potential as a SOTA solution in radiology report generation.

\subsubsection{Ablation Study}
\begin{figure}[h]
\centering
  \includegraphics[width=1\linewidth]{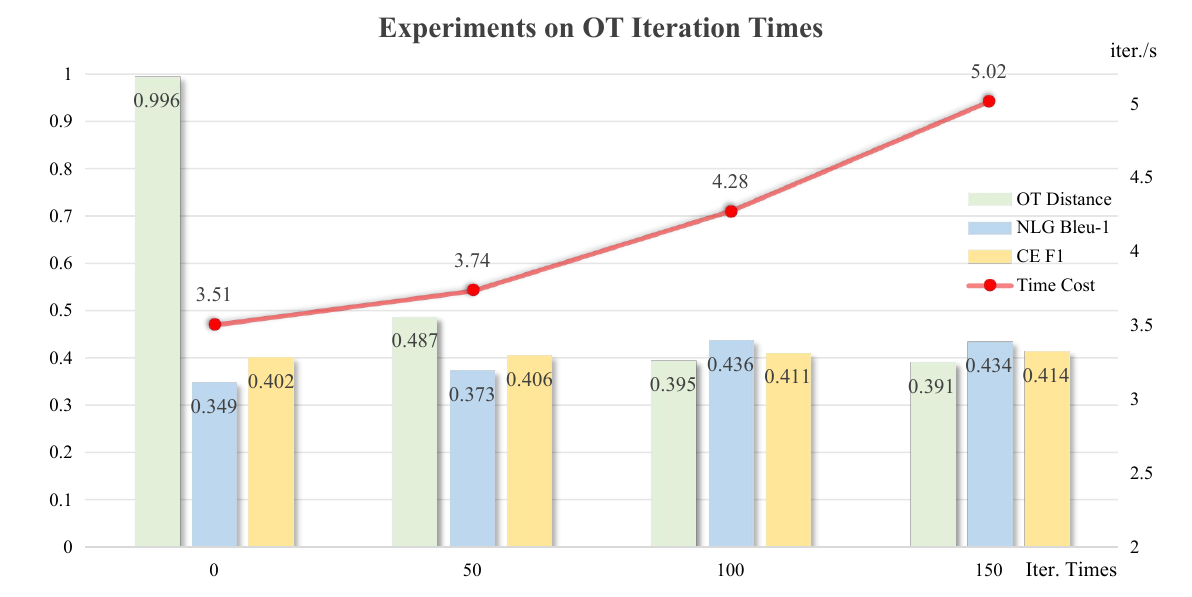}
\caption{Ablation Experiments on OT iteration times with MIMIC-CXR dataset.}
\label{fig:zhexian}   
\end{figure}

\begin{table}[t]
\centering
\small
\setlength{\tabcolsep}{0.9mm}
    \renewcommand{\arraystretch}{1}{
\caption{\small Ablation study of each module in OTDRG on MIMIC-CXR dataset. In this table, IRCP represents contrastive pre-training, and the DLE denotes the usage of the disease labels extraction module. The best results are in \textbf{bold}. Pre. and Rec. are the abbreviations of Precision and Recall, respectively. In NLG metrics, MET. represents METEOR and RO.-L represents ROUGE-L.}
\label{tab:results_ablation}
\begin{tabular}{lllccccccc}
\toprule	 
\multirow{2}{*}{IRCP} & \multirow{2}{*}{DLE} & \multirow{2}{*}{FIOT} & \multicolumn{3}{c}{\textbf{CE metrics}} & \multicolumn{4}{c}{\textbf{NLG metrics}}\\ \cmidrule(r){4-6} \cmidrule(r){7-10}
 & & &Pre. & Rec. & F1 & BLEU-1 & BLEU-4 & MET. & RO.-L \\
\midrule
 &  &  & 0.310 & 0.266 & 0.263 & 0.384 & 0.113 & 0.148 & 0.271 \\
\hspace{2mm}\cmark &  &  & 0.344 & 0.292 & 0.287 & 0.387 & 0.116 & 0.157 & 0.277 \\
 & \hspace{2mm}\cmark &  & 0.379 & 0.346 & 0.337 & 0.399 & 0.123 & 0.162 & 0.281 \\
\hspace{2mm}\cmark & \hspace{2mm}\cmark &  & 0.424 & 0.326 & 0.349 & 0.402 & 0.127 & 0.163 & \textbf{0.301} \\
\hspace{2mm}\cmark & \hspace{2mm}\cmark & \hspace{2mm}\cmark & \textbf{0.474} & \textbf{0.448} & \textbf{0.436} & \textbf{0.411} & \textbf{0.128} & \textbf{0.164} & 0.295 \\  
\bottomrule
\end{tabular}
}
\end{table}

\begin{figure*}[h]
\centering
  \includegraphics[width=1\linewidth]{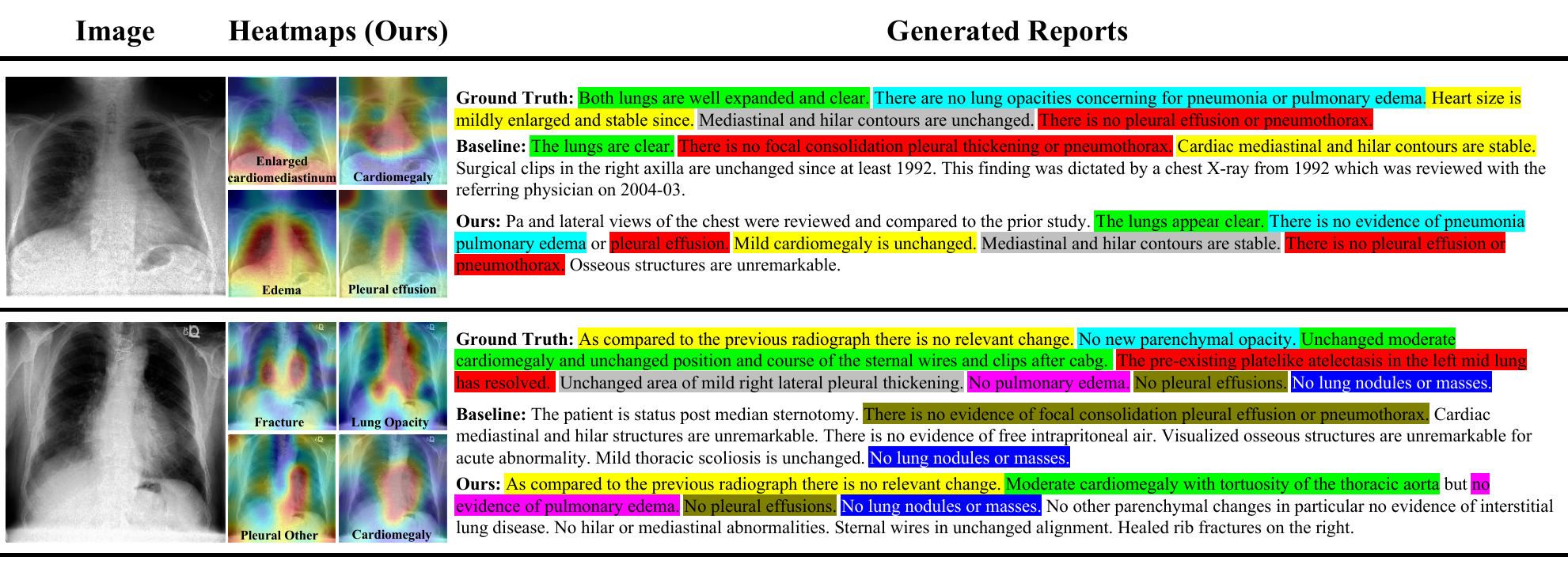}
 \caption{Visualization of the baseline model and our model. Highlights in different colors correspond to different diagnoses in the original report. Heatmaps visualize the transport plan $P^*$ on disease labels.}
\label{fig:vis}   
\end{figure*}

\begin{table}[h]
\caption{\small Ablation Experiments on entropic regulation hyper-parameter \textbf{$\epsilon$} of Sinkhorn-Knopp algorithm on MIMIC-CXR dataset.}
\label{tab: ablation_epsilon}
\centering
\begin{tabular}{lccc}
\toprule	 
{\textbf{$\varepsilon$}} & BLEU-1 & BLEU-4 & CE F1 \\
\midrule
0.01 & 0.365 & 0.107 & 0.391 \\
0.05 & 0.391 & 0.118 & 0.423 \\
0.10 & 0.411 & 0.128 & 0.436 \\
0.12 & 0.401 & 0.115 & 0.411 \\
\bottomrule
\end{tabular}
\end{table}
To validate the effectiveness of our proposed module for the radiology report generation task, especially in clinical quality, we conduct ablation experiments on the MIMIC-CXR dataset, as shown in Table~\ref{tab:results_ablation}. 
First, we test the baseline model on CE metrics and found that conventional methods for generating radiology reports with LLMs perform poorly on clinical indicators, where the harmonic mean F1 score in CE metrics was only 0.263.
Through the IRCP module, the visual encoder is more powerful during OTDRG training, and the model's performance is improved on both clinical-wise and NLG-wise.
Next, by employing the DLE module, we find that all three CE metrics had noticeable enhancement, and the Precision score increased from 0.310 to 0.379, and the Recall score increased from 0.266 to 0.346. 
However, relatively speaking, although we have enhanced the quality of the generated reports from a medical standpoint, this remains insufficient to meet the stringent requirements for clinical applicability.
So we employ both IRCP module and DLE module for further evaluation, and the result shows that the cooperation of the two modules works incredibly well on all CE scores, the precision score reached 0.424, and the F1 score increased from 0.263 to 0.349, which means that we have surpassed many of the previously proposed methods just with IRCP module and DLE module.
Our FIOT module aims to bridge the modal gap between the image modal and text modal, so the FIOT must follow the usage of the DLE.
After incorporating the FIOT module, our approach achieves a significant improvement in CE metrics, where Precision and Recall reached 0.474 and 0.436, greatly enhancing the clinical quality of LLM-generated radiology reports.
Although the improvement in NLG metrics is not as pronounced as in CE, our method still shows a notable enhancement compared to the baseline.

We also conduct ablation studies to further explore the impact of the number of OT iterations and the value of $\varepsilon$ in the Sinkhorn-Knopp algorithm.
As depicted in Figure~\ref{fig:zhexian}, the model's performance demonstrates a roughly positive correlation with the number of iterations, particularly in terms of the CE metrics. 
This is primarily because the number of OT iterations greatly affects the accuracy of the optimal transport plan calculation. 
Consequently, the accuracy of the OT results significantly impacts LLMs in generating diagnosis reports.
It should be noted that too many OT iterations will not enhance the performance of the model, but will increase the training time.
Table~\ref{tab: ablation_epsilon} indicates that $\varepsilon$ influences the quality of generated results, which further proves of the effectiveness of the FIOT module. In our experiments, we choose $0.10$ as the value of $\varepsilon$.

\subsection{Visualization}
In Figure~\ref{fig:vis}, we demonstrate the superiority of OTDRG on the radiology report generation task by presenting the visualization. 
Compared to the ground truth diagnostic report, OTDRG can accurately identify the cases in the images comprehensively, and its sentence construction is more similar to the ground truth, closer to the way real doctors express themselves when diagnosing cases. 
In contrast, the baseline method can only identify some cases, and due to the specialty of text generation by LLMs, it produces sentences that are irrelevant to the images awaiting diagnosis.
This phenomenon may be caused by the LLMs hallucination, where the model cannot learn high-quality medical knowledge from low-quality raw inputs, resulting in low-quality or even erroneous information being generated in the report.
However, our method provides higher-quality data that bridges the gap between image and text modals, which makes the model maximize the potential of LLMs for text generation, enabling them to better understand the input images and produce accurate diagnostic reports. Notably, the heatmaps of some disease labels are also displayed in Figure~\ref{fig:vis}, and they clearly visualize the transportation plan $P^*$. The results show that the FIOT module correctly focuses on the disease-related areas in the X-ray image, which also accurately corresponds to the generated reports, further demonstrating the effectiveness of our approach.

\section{Conclusion}

In this work, we propose the OTDRG model, a new framework that utilizes Large Language Models (LLMs) to generate radiological reports, addresses modal alignment problems by introducing OT (Alignment transformation), and unlocks the potential of LLMs through comprehensive model design. Compared to SOTA, the method achieved significant improvements on both NLG (natural language generation) and CE (clinical efficacy) metrics on MIME-CXR and IU X-Ray datasets. The ablation study further validates the contribution of each module in our model, particularly the FIOT module, which plays a key role in bridging the modal gap between medical images and text. Overall, OTDRG's success lies in its ability to effectively model the correspondence between X-ray images and disease labels, providing high-quality input to LLMs to generate linguistically coherent and clinically accurate reports.


\bibliographystyle{ACM-Reference-Format}
\bibliography{sample-base}

\end{document}